\documentclass[10pt,twocolumn,letterpaper]{article}

\usepackage{iccv}
\usepackage{times}
\usepackage{epsfig}
\usepackage{graphicx}
\usepackage{amsmath}
\usepackage{amssymb}
\usepackage{multirow}
\usepackage{adjustbox}
\usepackage[table]{xcolor}
\usepackage{pgfplots}
\pgfplotsset{width=7cm,compat=1.14}
\usepackage[linesnumbered,ruled]{algorithm2e}

\usepackage[pagebackref=true,breaklinks=true,letterpaper=true,colorlinks,bookmarks=false]{hyperref}

\iccvfinalcopy 

\def\iccvPaperID{****} 

\ificcvfinal\fi

\begin{document}

\title{Evaluating Menu OCR and Translation: A Benchmark for Aligning Human and Automated Evaluations in Large Vision-Language Models}

\author{
   Zhanglin Wu$^{1}$, Tengfei Song$^{1}$, Ning Xie$^{1}$, Mengli Zhu$^{1}$,  Weidong Zhang$^{1}$,\\
   Shuang Wu$^{1}$, Pengfei Li$^{1}$, Chong Li$^{1}$, Junhao Zhu$^{1}$, Hao Yang$^{1}$, Shiliang Sun$^{2}$\\
  $^{1}$Huawei Translation Service Center, Nanjing, China\\
  $^{2}$Shanghai Jiao Tong University, Shanghai, China\\
  \tt $^{1}$\{wuzhanglin2, songtengfei2, nicolas.xie, zhumengli,  zhangweidong17,\\
  \tt wushuang42, lipengfei203, august.li, zhujunhao, yanghao30\}@huawei.com \\
  \tt $^{2}$shiliangsun@gmail.com \\
  }
  

\maketitle
\ificcvfinal\thispagestyle{empty}\fi

\begin{abstract}



The rapid advancement of large vision-language models (LVLMs) has significantly propelled applications in document understanding, particularly in optical character recognition (OCR) and multilingual translation. However, current evaluations of LVLMs, like the widely used OCRBench, mainly focus on verifying the correctness of their short-text responses and long-text responses with simple layout, while the evaluation of their ability to understand long texts with complex layout design is highly significant but largely overlooked. 
In this paper, we propose Menu OCR and Translation Benchmark (MOTBench), a specialized evaluation framework emphasizing the pivotal role of menu translation in cross-cultural communication. MOTBench requires LVLMs to accurately recognize and translate each dish, along with its price and unit items on a menu, providing a comprehensive assessment of their visual understanding and language processing capabilities. Our benchmark is comprised of a collection of Chinese and English menus, characterized by intricate layouts, a variety of fonts, and culturally specific elements across different languages, along with precise human annotations. 
Experiments show that our automatic evaluation results are highly consistent with professional human evaluation.
We evaluate a range of publicly available state-of-the-art LVLMs, and through analyzing their output to identify the strengths and weaknesses in their performance, offering valuable insights to guide future advancements in LVLM development. MOTBench is available at \url{https://github.com/gitwzl/MOTBench}. 
\end{abstract}

\section{Introduction}

\begin{figure}[htbp]
\begin{center}
\includegraphics[width=0.47\textwidth, height=0.5\textwidth]{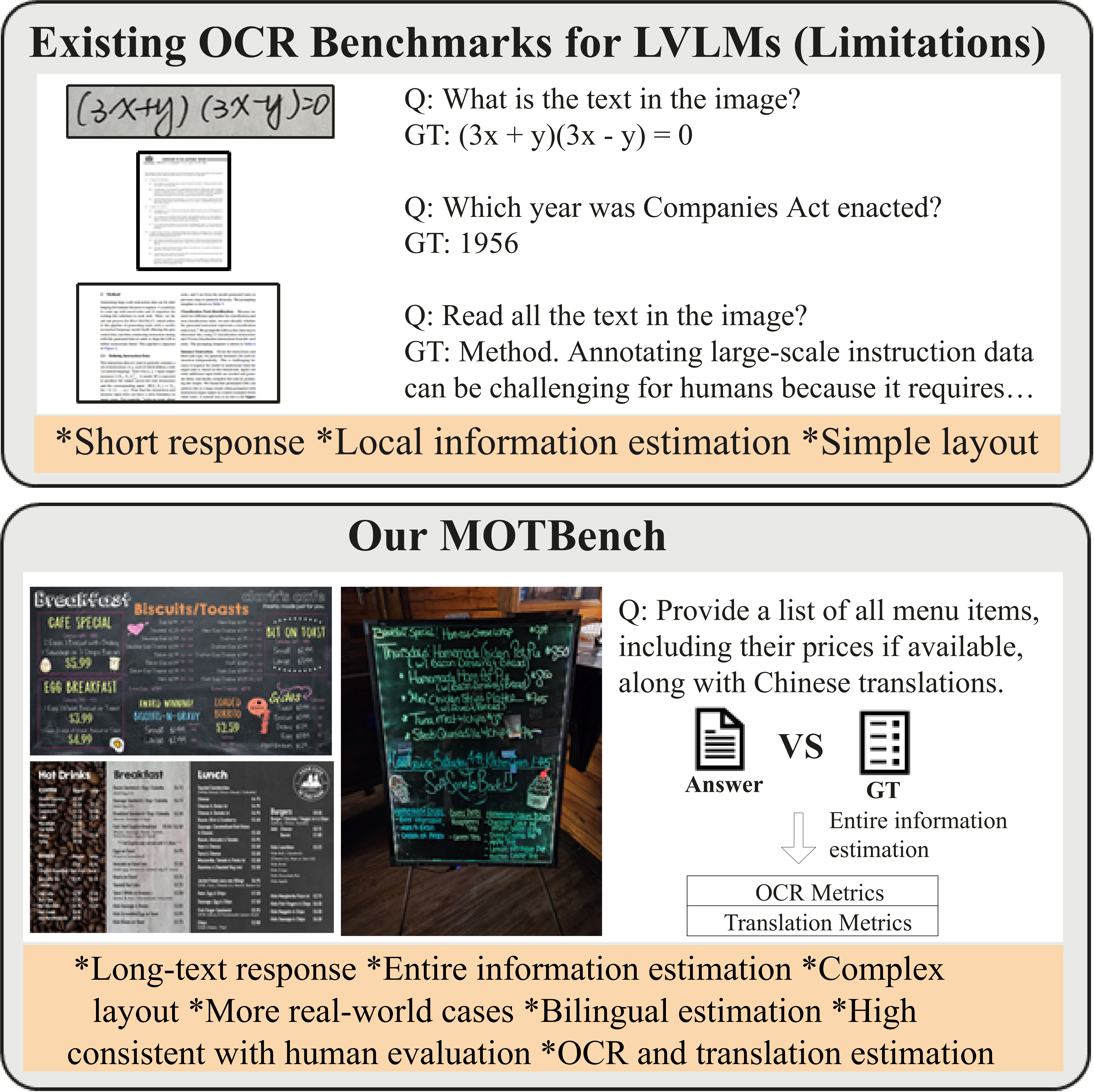}
\caption{An illustration comparing existing OCR benchmarks with our MOTBench.}\label{fig:intro}
\end{center}
\vspace{-0.5cm}
\end{figure}

Translating menu images \cite{hu2024bridging} is a significant technology for overcoming language barriers, empowering travelers and non-native speakers to comprehend local food options and make well-informed meal choices. However, conventional Optical Character Recognition (OCR) methods \cite{nguyen2021survey} frequently underperform in practical scenarios, particularly when dealing with menus that have complex layouts, handwritten text, or decorative fonts, often resulting in suboptimal translation outputs.


 Large Vision-Language Models (LVLMs) \cite{alayrac2022flamingo,driess2023palm,huang2023language,li2023blip} expand the potential of artificial intelligence to integrate textual and visual understanding, enabling significant advancements in OCR and translation tasks.
LVLMs such as GPT-4V \cite{2023GPT4VisionSC}, GPT-4o \cite{hurst2024gpt}, Deepseek\cite{lu2024deepseek}, Gemini-Pro-V \cite{team2023gemini}, along with some open-source models like MiniCPM \cite{yao2024minicpm} and InternVL \cite{wang2024enhancing}, have made significant progress in OCR and translation. 
Simultaneously, a variety of benchmarks have been developed to evaluate the various capabilities of LVLMs, such as MMBench \cite{liu2025mmbench}, MMMU \cite{yue2024mmmu}, MME \cite{fu2024mmecomprehensiveevaluationbenchmark}, and OCRBench \cite{liu2023hidden}. 


However, these benchmarks often focus on question-answering based on local information or overly rely on judgment-based or selection-based evaluation modes, failing to fully capture the complex challenges in real-world scenarios. These benchmarks may even obscure the strengths and weaknesses of LVLMs, hindering meaningful progress. Designing task-specific evaluation measures that capture the nuances of real-world challenges, particularly in structured, context-rich data such as menus and ensuring their alignment with human evaluation still remains a challenging issue. 


To bridge these gaps and establish a more reliable approach for evaluating the long-text comprehension and translation capabilities of LVLMs, we introduce a \textit{Menu
OCR and Translation Benchmark (MOTBench)} designed to evaluate their performance on Menu OCR and translation tasks.
As illustrated in Figure \ref{fig:intro}, MOTBench emphasizes handling long-text responses, comprehensively extracting entire information, and tackling complex layouts, making it better suited for real-world scenarios. In contrast, existing OCR benchmarks primarily focus on short responses, localized information estimation, and simple layouts.



LVLMs frequently produce a wide variety of output formats, including structured text and free-form translations, which complicates the evaluation process and makes it difficult to accurately pinpoint differences between the outputs of LVLMs and the ground truth through direct matching. Comparing the predicted menu directly with the ground truth often leads to confusion, resulting in significant discrepancies with human evaluation. To improve evaluation accuracy, we propose a fine-grained, pairwise comparison method that evaluates each menu item individually in tasks of OCR and translation. For OCR, we employ character-level matching to ensure precise recognition.  
For the translation task, we extract the translation results for each dish, and then use commonly used evaluation metrics from previous work \cite{wei2023text,guo2024novel}, such as BLEU \cite{papineni2002bleu} and COMET \cite{rei2022comet}, to calculate the similarity to the reference translations. Ultimately, our automatic evaluation results demonstrate strong alignment with human evaluation, ensuring the reliability of our automated assessment.


The key contributions of this paper are as follows:
\begin{itemize} 
\item We release a high-quality Menu OCR and Translation Benchmark, which covers both English and Chinese menus with carefully curated annotations from professional translators. It addresses the real-world complexities of menu layouts and text styles, serving as a robust dataset for advancing the capabilities of LVLMs.
\vspace{-0.2cm}
\item We propose pairwise evaluation protocol tailored for long-text OCR and translation tasks, allowing for granular assessment of complex menu structures that effectively addresses the challenges of estimating long-text OCR and dish translation, with results highly consistent with human evaluation.
\vspace{-0.2cm}
\item We evaluate 17 open-source LVLMs alongside 7 closed-source LVLMs using our MOTBench, providing a benchmark to drive progress in visual language technologies and expand their practical applications.
\end{itemize}


\begin{figure*}[htbp]
\begin{center}
\includegraphics[width=0.95\textwidth, height=0.58\textwidth]{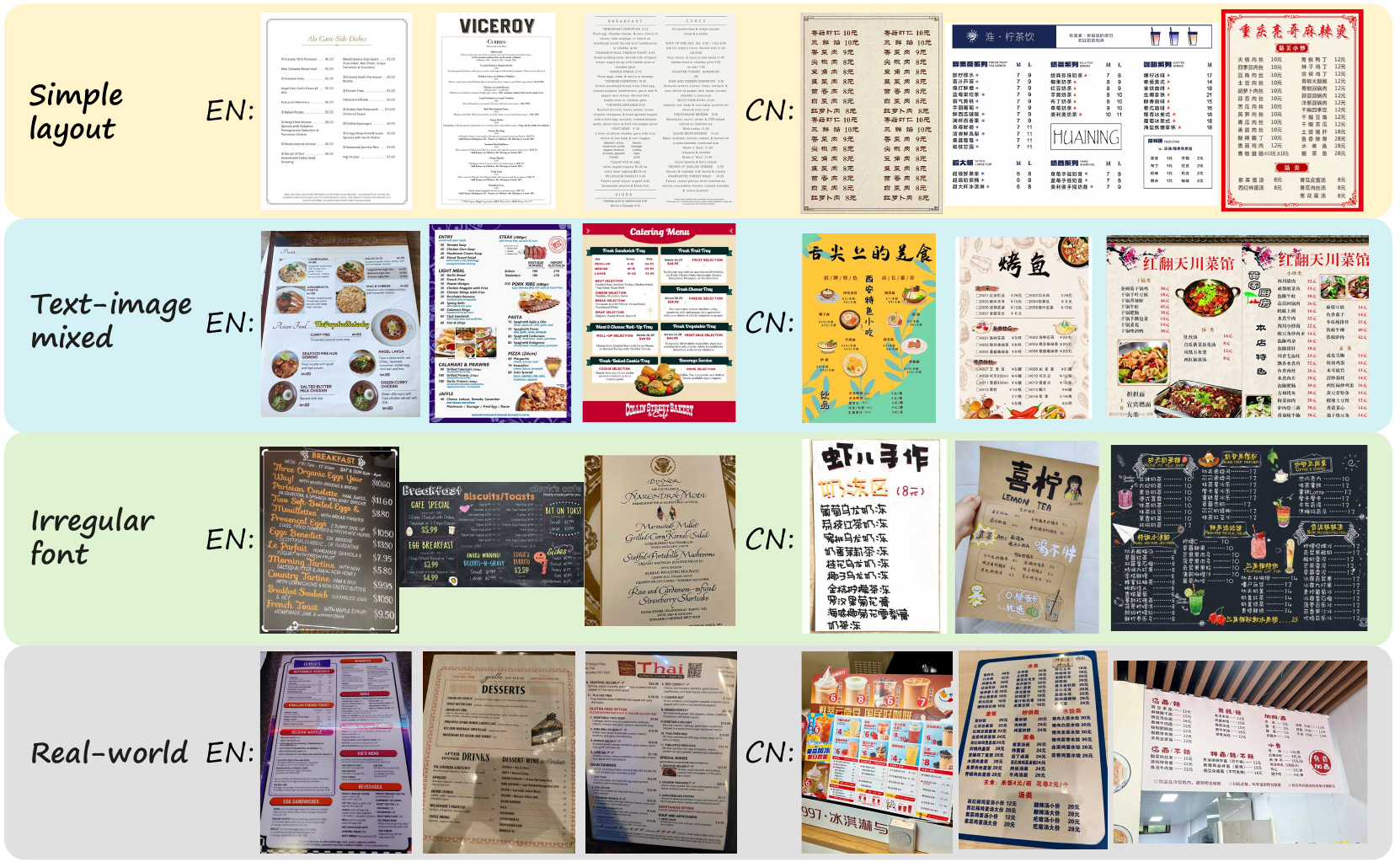}
\caption{Examples sampled from MOTBench are shown. The menu images are categorized into four groups: 'Simple Layout,' 'Text-Image Mixed,' 'Irregular Font,' and 'Real-World'. Both English and Chinese menus are included. For each menu image, the ground truth annotations include all dish items, their prices, and their corresponding translations. We employ two distinct prompts to evaluate OCR capability and translation capability separately.}\label{fig:menu}
\end{center}
\vspace{-0.8cm}
\end{figure*}

\section{Related Work}




The rise of LVLMs not only drives the deep integration of computer vision and natural language processing but also brings new possibilities to applications such as intelligent interaction and content generation. However, how to scientifically evaluate the performance of these models and further promote their optimization and innovation remains a key concern for both academia and industry. Next, we explore the evolution of menu translation, the development history of LVLMs, and the latest advancements in their evaluation benchmarks to reveal the current technical challenges and future development directions.

\subsection{Menu Translation}


In the field of menu translation, it is crucial to handle elements that hold unique and significant meanings within specific cultures. As pointed out by \cite{aixela1996culture}, these elements are common in the source culture but may be uncommon in the target culture, thus requiring specific translation strategies. Pallatt et al. \cite{pellatt2010thinking} specifically studied the translation of Chinese menus, emphasizing the importance of adaptation strategy. Chou et al. \cite{chou2016translational} proposed a neutralization strategy aimed at finding a balance between foreignization and domestication to ensure that the translation is faithful to the original meaning and aligns with the expressive habits of the target culture. Amenador et al. \cite{amenador2022translation} through their analysis of Chinese dish name translations, found that detailed explanations are a common method used by human translators. Zhang et al. \cite{zhang2024cultural} drew on these strategies and applied them to zero-shot prompting, thereby enhancing the performance of large language models (LLMs) in menu translation. However, there is still a lack of in-depth exploration into the application of LVLMs in menu translation. 
Thus, we introduce a high-quality Menu OCR and Translation Benchmark. This benchmark covers both English and Chinese menus, capturing the real-world complexities of menu layouts and text styles. It serves as a robust dataset to advance the capabilities of large visual language models.
\subsection{LVLMs}

The development of LVLMs is closely linked to the rise of LLMs \cite{chung2024scaling,waghmare2023introduction,touvron2023llama}. LLMs have demonstrated powerful capabilities in text understanding and generation through large-scale pre-training and fine-tuning. LVLMs build on this foundation by extending these abilities to span both vision and language modalities, enabling the handling of complex multi-modal tasks. 


BLIP-2 \cite{li2023blip} introduced the Querying Transformer \cite{zhao2022queryformer} to effectively bridge the gap between vision and language models. Flamingo \cite{alayrac2022flamingo} and OpenFlamingo \cite{awadalla2023openflamingo} improved on this approach by integrating frozen pretrained LLMs \cite{mosaicml7b,wei2024towards} with gated cross-attention dense layers \cite{jaegle2021perceiver} to better handle visual inputs.
LLaVA \cite{li2024llava} marked a milestone as the first model to utilize GPT-4 \cite{hurst2024gpt} for generating multimodal instruction-following data, while MiniGPT-4 \cite{zhu2023minigpt} focused on aligning visual modules with LLMs to enhance multimodal understanding. Similarly, mPLUG-Owl \cite{ye2023mplug} and mPLUG-Owl2 \cite{ye2024mplug} emphasized stronger collaboration between image and text modalities. Building on LLaVA, LLaVAR \cite{zhang2023llavar} improved OCR capabilities by training on diverse text datasets and using a higher-resolution CLIP model \cite{radford2021learning} as the vision encoder. However, the aforementioned methods are not capable of handling images with complex layouts, which limits their widespread application.

Recent advancements include LLaVA-OneVision \cite{li2024llava}, which is trained on extensive, high-quality datasets and excels across single-image, multi-image, and video scenarios. 
MiniCPM-V 2.6\cite{yao2024minicpm}, based on SigLip-400M and Qwen2-7B, supports diverse visual modalities while reducing visual token counts to boost efficiency. 
InternVL2.5-MPO \cite{wang2024enhancing} introduces preference optimization to enhance the multimodal reasoning capabilities of MLLMs. 
DeepSeek-VL2 \cite{deepseek-vl2} introduces an advanced series of large Mixture-of-Experts (MoE) LVLMs with enhanced capabilities in visual question answering, optical character recognition, document/table/chart understanding, and visual grounding. 
However, these models still face difficulties in handling images with long text and complex layouts. 
Besides the open-source models, the closed-source models like GPT-4V \cite{2023GPT4VisionSC}, GPT-4o \cite{hurst2024gpt}, kimi \cite{KimiVision}, and Gemini-Pro-V \cite{team2023gemini} demonstrate outstanding performance across a wide range of visual tasks, solidifying the transformative impact of LVLMs on AI capabilities. How to measure their text perception capabilities and achieve results consistent with human evaluation remains a challenging problem.


\subsection{Benchmarks for LVLMs}



LVLMs have shown impressive capabilities in handling various vision-language tasks, highlighting the limitations of traditional single-task benchmarks \cite{antol2015vqa,hudson2019gqa,krishna2017visual,marino2019ok} in providing a comprehensive evaluation of their performance. In response, current evaluation benchmarks aim to provide a relatively comprehensive assessment of LVLMs' multimodal reasoning ability. For example, MMBench \cite{liu2025mmbench} and SEED-Bench-2 \cite{li2024seed} evaluate LVLMs through multiple-choice questions across different dimensions, while MMTBench \cite{ying2024mmt} and MMMU \cite{yue2024mmmu} focus on knowledge-based, domain-specific multiple-choice questions for LVLMs evaluation. Additionally, MME \cite{fu2024mmecomprehensiveevaluationbenchmark} evaluates LVLMs through manually designed judgment questions. However, the over-reliance on multiple-choice and judgment-based evaluation methods limits the detailed insight into LVLMs, making it difficult to conduct in-depth analysis of their capabilities.

To mitigate the issues arising from such dependencies, MULTI \cite{zhu2024multi} combines multiple-choice and short-answer questions, offering a more granular evaluation of LVLMs' ability to understand complex tables, images, and long-context reasoning. This also drives the shift in LVLMs evaluation scheme from multiple-choice and judgment questions to short-answer questions. Furthermore, OCRBench \cite{liu2023hidden} evaluates LVLMs' reasoning abilities with a carefully selected set of 1,000 question-answer pairs, entirely removing the reliance on judgment and multiple-choice questions. However, these evaluation benchmarks focus on verifying the correctness of LVLMs' generation of short text responses and are not yet sufficient to capture the complex challenges present in real-world scenarios. In contrast, our evaluation benchmark focuses on evaluating the correctness of LVLMs in generating document-level long texts, providing a more comprehensive and fine-grained evaluation of their multimodal reasoning capabilities.


\section{MOTBench}


We introduce MOTBench, which aims to provide a scientific and comprehensive evaluation framework for evaluating the performance of LVLMs in menu OCR and translation tasks through systematic evaluation dataset, evaluation strategy, and instruction design. Next, we elaborate on the construction principles of MOTBench.

\subsection{Evaluation Dataset}

\begin{table}
\begin{center}
\begin{adjustbox}{width=0.9\columnwidth,center}
\begin{tabular}{lcccc}
\hline
\multirow{2}{*}{Menu type} & \multicolumn{2}{c}{Chinese} & \multicolumn{2}{c}{Enlgish} \\
  & Images & Items & Images & Items \\
\hline
Simple layout & 20 & 751 & 20 & 418 \\
Text-image mixed & 20 & 701 & 20 & 535 \\
Irregular font & 10 & 202 & 10 & 155 \\
Real-world & 15 & 674 & 15 & 359 \\
\hline
All & 65 & 2328 & 65 & 1467\\
\hline
\end{tabular}
\end{adjustbox}
\end{center}
\caption{The distribution of evaluation dataset across different languages and menu types.}\label{evadata}
\vspace{-0.4cm}
\end{table}


We collect publicly available Chinese and English menu images from the Internet to build the evaluation dataset required for MOTBench. First, we categorize the menu images into four types based on complexity: simple layout, text-image mixed, irregular font, and real-world. Then, we select representative menu images from each type to form the evaluation dataset. Some menu images are shown in Figure \ref{fig:menu}. Finally, we invite language experts to annotate the dish items in each menu image with corresponding Chinese and English text. As shown in Table \ref{evadata}, we select 65 menu images for each language, totaling 2,328 Chinese dish items and 1,467 English dish items.


\begin{figure}[t]
\begin{center}
\includegraphics[width=0.44\textwidth, height=0.15\textwidth]{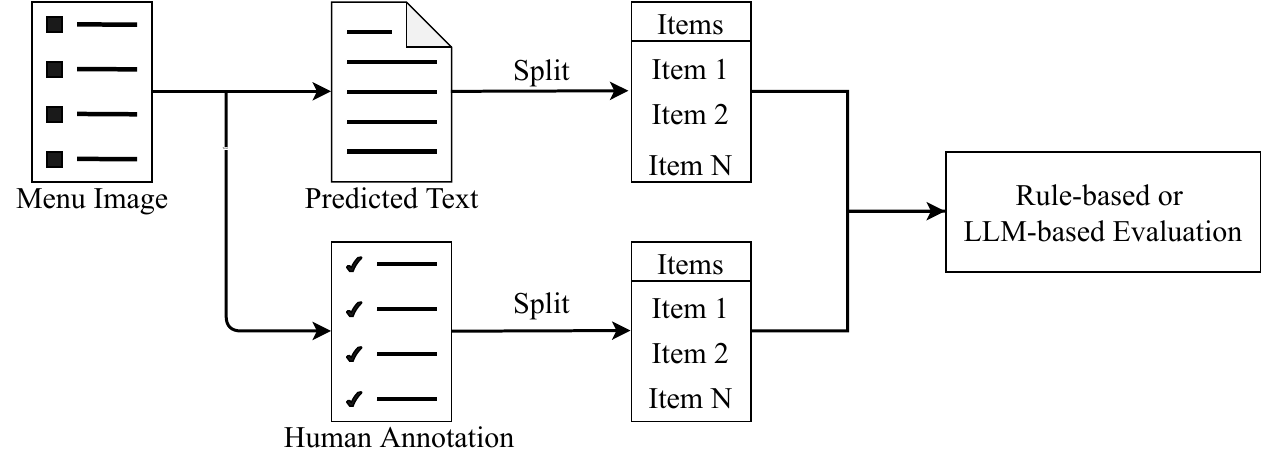}
\caption{The illustration of one-by-one comparison menu evaluation strategy in our MOTBench.}\label{fig:evaluation_example}
\end{center}
\vspace{-0.4cm}
\end{figure}

\subsection{Evaluation Strategy}


The menu images in the MOTBench evaluation dataset are rich in content and feature a wide variety of dishes. Therefore, we adopt a one-by-one menu comparison evaluation strategy to evaluate the performance of LVLMs in both menu OCR and translation tasks, the illustration of this process is shown in Figure \ref{fig:evaluation_example}.

\paragraph{Menu OCR Evaluation.}


In each menu image of the MOTBench evaluation dataset, there are many dishes, and most of the dishes also come with prices and unit items. For the menu OCR task, we design a coarse-grained evaluation metric and a fine-grained evaluation metric. As shown in Algorithm \ref{alg1}, we use both dish recognition accuracy $acc_1$ and dish and its associated item recognition accuracy $acc_2$ as evaluation metrics for the menu OCR task.

\IncMargin{0em}
\begin{algorithm} \SetKwInOut{Input}{Input}\SetKwInOut{Output}{Output}

    \Input{OCR results $V$ for $n$ images, $m$ dish annotation results $D$ and their associated item annotation results $I$ for each image.}
    \Output{$acc_1$ and $acc_2$}
    $A_1=A_2=[]$\;
    \For{$i\leftarrow 1$ \KwTo $n$}{
        \For{$j\leftarrow 1$ \KwTo $m$}{
            $ correct_1=correct_2=0$\;
            \If{$D_{i,j}$ in $V_i$}
               {$correct_1=1$\;
               \If{$I_{i,j}$ in $V_i$}
               {$correct_2=1$\;}
               }
            $A_1.append(correct_1)$\;
            $A_2.append(correct_2)$\;
        }
    }
    $acc_1=\frac{sum(A_1)}{len(A_1)}*100\%$\; 
    $acc_2=\frac{sum(A_2)}{len(A_2)}*100\%$\;
    
\caption{OCR evaluation process}\label{alg1}
\end{algorithm}
\DecMargin{0em}

\paragraph{Menu Translation Evaluation.}


In the menu translation task, we use BLEU \footnote{\url{https://github.com/mjpost/sacrebleu}} and COMET \footnote{\url{https://huggingface.co/Unbabel/wmt22-comet-da}}, which are commonly used evaluation metrics for text translation tasks, to evaluate the translation quality of LVLMs. Due to the significant differences between the output format of LVLMs and the standard answers, directly using these two evaluation metrics becomes challenging. Therefore, we introduce an LLM to extract the translation results for each dish, and then calculate the BLEU and COMET scores between the translation results and the reference translations. As shown in Algorithm \ref{alg2}, when extracting translation results, we also incorporate some rules to reduce unnecessary LLM calls in order to minimize false positives.

\IncMargin{0em}
\begin{algorithm} \SetKwInOut{Input}{Input}\SetKwInOut{Output}{Output}

    \Input{Translation results $V$ for $n$ images, $m$ dish annotation results $D$ and their reference translations $R$ for each image.}
    \Output{$bleu$ and $comet$}
    $bleus=comets=[]$\;
    \For{$i\leftarrow 1$ \KwTo $n$}{
        \For{$j\leftarrow 1$ \KwTo $m$}{
            $mt=""$\;
            \If{$D_{i,j}$ in $V_i$}
               {$mt=llm\_extra\_mt(D_{i,j}, V_i)$\;
               \If{$mt$ not in $V_i$}
               {$mt=""$\;}
               }
            $bleus.append(cal\_bleu(mt, R_{i,j}))$\;
            $comets.append(cal\_comet(D_{i,j}, mt, R_{i,j}))$\;
               
        }
    }
    $bleu=\frac{sum(bleus)}{len(bleus)}*100\%$\;
    $comet=\frac{sum(comets)}{len(comets)}*100\%$\;
    
\caption{Translation evaluation process}\label{alg2}
\end{algorithm}
\DecMargin{0em}


\subsection{Instruction Design}


When using LVLMs to generate OCR results for menu images, our designed instruction is: “$<$image$>$$\backslash$nProvide a list of all menu items, including their prices if available.”

When using LVLMs to generate English/Chinese translation results for menu images, our designed instruction is: “$<$image$>$$\backslash$nPlease provide all menu items in a bilingual format (English and Chinese). Each dish should be listed on a separate line, with the English and Chinese descriptions for the same dish appearing on the same line. If a dish includes additional information such as price or unit of measurement, ensure that this information is placed on the same line as the dish description.”

When extracting the translated dish names from the output of LVLMs using LLM, we choose LLama-3.1-8B-Instruct\footnote{\url{https://huggingface.co/meta-llama/Llama-3.1-8B-Instruct}\label{llama3.1}} and design the extraction instruction as: "Given the text \{lvlm\_translation\_result\}, find the English/Chinese translation for the specified word \{src\_dish\}. Only return the translation without any additional information. Do not generate translations by yourself. If no translation is found, return empty string."


\begin{table*}[ht]
\begin{center}
\begin{adjustbox}{width=1.45\columnwidth,height=3.7cm,center}
\begin{tabular}{lcccccccccc}
\hline
\multirow{2}{*}{EN} & \multicolumn{5}{c}{$acc_1$} & \multicolumn{5}{c}{$acc_2$} \\
& \textbf{S} & \textbf{T} & \textbf{I} & \textbf{R} & \textbf{All} & \textbf{S} & \textbf{T} & \textbf{I} & \textbf{R} & \textbf{All} \\
\hline
\rowcolor{gray!20} \multicolumn{11}{c}{Open-source LVLMs}\\
BLIP2-Opt-6.7B & 0.00 & 0.00 & 1.29 & 0.00 & 0.14 & 0.00 & 0.00 & 0.00 & 0.00 & 0.00 \\
InstructBLIP-Vicuna-7B & 2.39 & 0.75 & 9.03 & 6.13 & 3.41 & 0.00 & 0.00 & 1.94 & 0.56 & 0.34 \\
mPLUG-Owl3-7B-241101 & 12.20 & 9.72 & 15.48 & 11.70 & 11.52 & 2.87 & 2.62 & 1.94 & 1.67 & 2.39 \\
LLaVA1.5-7B & 18.66 & 12.15 & 18.06 & 18.11 & 16.09 & 1.20 & 2.43 & 4.52 & 1.95 & 2.18 \\
LLaVA1.5-13B & 25.12 & 22.43 & 30.32 & 25.91 & 24.88 & 3.11 & 3.55 & 3.87 & 0.28 & 2.66 \\
LLaVA-Next-Interleave-Qwen-7B & 39.23 & 34.02 & 44.52 & 47.08 & 39.81 & 16.03 & 14.39 & 22.58 & 10.86 & 14.86 \\
LLaVA-OneVision-Qwen2-7B & 76.56 & 70.09 & 74.19 & 69.08 & 72.12 & 19.14 & 40.37 & 27.74 & 27.58 & 29.86 \\
DeepSeek-VL2-Small-16B & 23.92 & 16.64 & 38.06 & 25.35 & 23.11 & 9.09 & 7.48 & 24.52 & 6.69 & 9.54 \\
DeepSeek-VL2-27B & 22.01 & 16.07 & 32.26 & 23.68 & 21.34 & 7.66 & 8.60 & 14.84 & 6.96 & 8.59 \\
Llama3.2-11B-Vision-Instruct & \textbf{95.93} & 89.72 & 77.42 & 71.03 & 85.62 & 29.90 & 30.09 & 36.77 & 32.59 & 31.36 \\
MiniCPM-Llama3-V2.5-8B & 80.38 & 78.69 & 68.39 & 77.16 & 77.71 & 48.56 & 47.48 & 38.06 & 26.18 & 41.58 \\
MiniCPM-V2.6-8B & 92.11 & 86.54 & 73.55 & 83.29 & 85.96 & 44.02 & 51.21 & 36.13 & 30.92 & 42.60 \\
Qwen2-VL-7B-Instruct & 86.84 & 73.27 & 84.52 & 74.65 & 78.66 & 45.45 & 52.15 & 65.16 & 39.00 & 48.40 \\
Qwen2.5-VL-7B-Instruct & 83.49 & 85.61 & \textbf{87.74} & 84.68 & 85.00 & 55.74 & \textbf{67.29} & \textbf{76.77} & 40.67 & 58.49 \\
InternVL2-8B & 88.76 & 82.62 & 78.06 & 73.82 & 81.73 & 45.22 & 56.07 & 52.26 & \textbf{51.25} & 51.40 \\
InternVL2.5-8B & 91.87 & 83.74 & 81.94 & \textbf{88.02} & 86.91 & 65.79 & 63.36 & 75.48 & 48.19 & 61.62 \\
InternVL2.5-8B-MPO & 94.50 & \textbf{90.09} & 85.16 & 87.47 & \textbf{90.18} & \textbf{72.01} & 64.49 & 73.55 & 50.14 & \textbf{64.08} \\
\hline
\rowcolor{gray!20} \multicolumn{11}{c}{Closed-source LVLMs}\\
GPT-4o-2024-11-20 & 98.09 & 94.77 & 89.68 & 90.53 & 94.14 & 81.58 & 86.17 & 79.35 & 71.87 & 80.64 \\
GPT-4.5-preview-2025-02-27 & \textbf{98.80} & 96.26 & \textbf{92.90} & \textbf{93.59} & \textbf{95.58} & 81.58 & 86.36 & 81.29 & \textbf{84.12} & 83.91 \\
Moonshot-V1-8K & 91.63 & 82.99 & 72.90 & 88.30 & 85.69 & 66.03 & 42.06 & 39.35 & 39.00 & 47.85 \\
Qwen-VL-Max-2024-11-19 & 94.98 & 94.58 & 78.06 & 89.69 & 91.75 & 78.95 & 87.66 & 72.26 & 69.64 & 79.14 \\
Qwen-VL-Max-2025-01-25 & 97.13 & 95.89 & 90.32 & 93.31 & 95.02 & 80.14 & 88.97 & 82.58 & 81.62 & 83.98 \\
Gemini-1.5-Flash & 97.13  & 93.46  & 92.26  & 91.36  & 93.87  & 79.43  & 85.05  & 81.94  & 74.37  & 80.50 \\
Gemini-2.0-Flash & 98.09 & \textbf{96.64} & \textbf{92.90} & 91.64 & 95.43 & \textbf{83.73} & \textbf{91.03} & \textbf{85.16} & 82.45 & \textbf{86.23} \\
\hline
\end{tabular}
\end{adjustbox}
\end{center}
\vspace{-0.2cm}
\caption{The $acc_1$ and $acc_2$ scores of the English menu OCR.}
\label{ocr_en_combined}
\end{table*}

\begin{table*}[ht]
\begin{center}
\begin{adjustbox}{width=1.45\columnwidth,height=3.7cm,center}
\begin{tabular}{lcccccccccc}
\hline
\multirow{2}{*}{CN} & \multicolumn{5}{c}{\textbf{$acc_1$}} & \multicolumn{5}{c}{\textbf{$acc_2$}} \\
 & \textbf{S} & \textbf{T} & \textbf{I} & \textbf{R} & \textbf{All} & \textbf{S} & \textbf{T} & \textbf{I} & \textbf{R} & \textbf{All} \\
\hline
\rowcolor{gray!20} \multicolumn{11}{c}{Open-source LVLMs}\\
BLIP2-Opt-6.7B & 0.00 & 0.00 & 0.00 & 0.00 & 0.00 & 0.00 & 0.00 & 0.00 & 0.00 & 0.00 \\
InstructBLIP-Vicuna-7B & 0.00 & 0.29 & 0.00 & 0.15 & 0.13 & 0.00 & 0.00 & 0.00 & 0.00 & 0.00 \\
mPLUG-Owl3-7B-241101 & 0.27 & 0.29 & 1.98 & 0.45 & 0.47 & 0.00 & 0.14 & 0.00 & 0.00 & 0.04 \\
LLaVA1.5-7B & 0.00 & 0.14 & 0.00 & 0.00 & 0.04 & 0.00 & 0.00 & 0.00 & 0.00 & 0.00 \\
LLaVA1.5-13B & 0.00 & 0.43 & 0.50 & 0.30 & 0.26 & 0.00 & 0.00 & 0.00 & 0.00 & 0.00 \\
LLaVA-Next-Interleave-Qwen-7B & 0.00 & 0.00 & 0.00 & 0.00 & 0.00 & 0.00 & 0.00 & 0.00 & 0.00 & 0.00 \\
LLaVA-OneVision-Qwen2-7B & 2.40 & 2.43 & 1.98 & 1.04 & 1.98 & 0.67 & 0.14 & 0.00 & 0.00 & 0.26 \\
DeepSeek-VL2-Small-16B & 11.19 & 16.12 & 5.94 & 2.37 & 9.66 & 6.92 & 5.71 & 4.46 & 0.89 & 4.60 \\
DeepSeek-VL2-27B & 11.45 & 16.98 & 3.47 & 3.41 & 10.09 & 7.59 & 7.99 & 0.99 & 1.19 & 5.28 \\
Llama3.2-11B-Vision-Instruct & 18.24 & 19.97 & 5.45 & 19.29 & 17.96 & 1.86 & 0.71 & 0.00 & 5.79 & 2.49 \\
MiniCPM-Llama3-V2.5-8B & 30.36 & 29.96 & 25.74 & 44.51 & 33.93 & 17.58 & 19.69 & 4.46 & 23.44 & 18.77 \\
MiniCPM-V2.6-8B & 49.27 & 62.91 & 46.04 & 61.13 & 56.53 & 32.49 & 52.21 & 20.79 & 39.91 & 39.56 \\
Qwen2-VL-7B-Instruct & 59.25 & 60.06 & 45.05 & 61.57 & 58.93 & 41.81 & 49.50 & 23.76 & 29.97 & 39.13 \\
Qwen2.5-VL-7B-Instruct & 71.37 & 65.05 & 47.03 & 77.15 & 69.03 & \textbf{62.98} & \textbf{60.77} & 42.08 & \textbf{72.11} & \textbf{63.14} \\
InternVL2-8B & 67.51 & 63.48 & 68.81 & 70.47 & 67.27 & 26.63 & 46.22 & 27.23 & 25.22 & 32.17 \\
InternVL2.5-8B & 64.45 & 58.06 & 78.71 & 80.42 & 68.38 & 56.19 & 51.50 & 58.42 & 57.86 & 55.46 \\
InternVL2.5-8B-MPO & \textbf{72.30} & \textbf{68.90} & \textbf{83.17} & \textbf{83.53} & \textbf{75.47} & 60.72 & 59.34 & \textbf{57.92} & 67.51 & 62.03 \\
\hline
\rowcolor{gray!20} \multicolumn{11}{c}{Closed-source LVLMs}\\
GPT-4o-2024-11-20 & 51.93 & 54.07 & 45.05 & 55.34 & 52.96 & 44.87 & 42.65 & 40.10 & 45.99 & 44.12 \\
GPT-4.5-preview-2025-02-27 & 69.64 & 72.75 & 45.54 & 76.26 & 70.40 & 58.59 & 62.91 & 35.15 & 64.09 & 59.45 \\
Moonshot-V1-8K & 81.36 & 83.74 & 82.18 & 89.47 & 84.49 & 61.38 & 59.91 & 39.11 & 63.20 & 59.54 \\
Qwen-VL-Max-2024-11-19 & 81.76 & 70.19 & 66.34 & 72.11 & 74.14 & 72.84 & 53.50 & 57.43 & 61.57 & 62.41 \\
Qwen-VL-Max-2025-01-25 & 80.69 & 80.46 & 60.89 & 88.58 & 81.19 & 72.70 & 66.62 & 50.99 & 81.75 & 71.61 \\
Gemini-1.5-Flash & 96.54  & 88.73  & 84.16  & 97.33  & 93.34  & 88.55  & 82.88  & 70.79  & 86.94  & 84.84 \\
Gemini-2.0-Flash & \textbf{97.60} & \textbf{93.01} & \textbf{91.09} & \textbf{98.37} & \textbf{95.88} & \textbf{90.41} & \textbf{90.87} & \textbf{75.74} & \textbf{90.06} & \textbf{89.18} \\
\hline
\end{tabular}
\end{adjustbox}
\end{center}
\vspace{-0.2cm}
\caption{The $acc_1$ and $acc_2$ scores of the Chinese menu OCR.}
\label{ocr_zh_combined}
\vspace{-0.5cm}
\end{table*}

\section{Experiments}


In this section, various open-source and closed-source LVLMs are evaluated on MOTBench, including BLIP2, InstructBLIP, mPLUG-Owl3, LLAVA, DeepSeek-VL2, Llama-3.2-Vision, MiniCPM-V, Qwen-VL2, Qwen-VL2.5, InternVL2, InternVL2.5, GPT-4o, GPT-4.5, Moonshot-V1, Qwen-VL-Max, and Gemini. Our evaluation is conducted under a zero-shot setting to evaluate the models' ability to generate accurate answers without fine-tuning. 


\subsection{Result}


\paragraph{English Menu OCR Evaluation Result.} The evaluation results of English (EN) menu OCR are presented in Tables \ref{ocr_en_combined}. There is a significant variation in OCR capabilities among different LVLMs. Some LVLMs, such as BLIP2 and InstructBLIP, essentially lack the ability to perform menu OCR. On the other hand, certain open-source LVLMs like Qwen2.5-VL-7B-Instruct and InternVL2.5-8B-MPO exhibit strong menu OCR capabilities. closed-source LVLMs, such as GPT-4.5-preview-2025-02-27 and Gemini-2.0-Flash, generally outperform most open-source models in menu OCR tasks. In terms of accuracy for recognizing only English dish names ($acc_1$), both open-source and closed-source LVLMs can achieve very high levels (InternVL2.5-8B-MPO's $acc_1$ is 90.18, and GPT-4.5-preview-2025-02-27's $acc_1$ is 95.58). However, there is still considerable room for improvement in the accuracy of recognizing entire English menu items ($acc_2$) for both open-source and closed-source LVLMs (InternVL2.5-8B-MPO's $acc_2$ is 64.08, and GPT-4.5-preview-2025-02-27's $acc_2$ is 86.23). Additionally, the accuracy of the same LVLM in recognizing dishes or entire menu items can vary depending on the type of English menu image. Most LVLMs tend to have slightly higher accuracy in recognizing simple layout (S) and text-image combined (T) menu images compared to those with irregular font (I) and real-world scenes (R).


\paragraph{Chinese Menu OCR Evaluation Result.} The evaluation results of Chinese (CN) menu OCR are presented in Tables \ref{ocr_zh_combined}. We observe a phenomenon that LVLMs, which perform strongly on English menu OCR, do not necessarily exhibit the same level of capability on Chinese menu OCR. For example, among open-source LVLMs, InternVL2.5-8B-MPO still has the strongest ability to recognize individual Chinese dishes, but its ability to recognize entire Chinese menu items is weaker than that of Qwen2.5-VL-7B-Instruct. Among closed-source LVLMs, GPT-4.5-preview-2025-02-27 shows a decline in its ability to recognize both individual Chinese dishes and entire Chinese menu items, performing significantly worse than Gemini-2.0-Flash.



Compared to the menu OCR task, the menu translation task poses a greater challenge to LVLMs. This is because LVLMs first need to utilize OCR capabilities to recognize the text information of the dishes in the images, and then use language processing abilities to perform the translation.

\begin{table*}[ht]
\begin{center}
\begin{adjustbox}{width=1.45\columnwidth,height=3.7cm,center}
\begin{tabular}{lcccccccccc}
\hline
\multirow{2}{*}{EN2CN} & \multicolumn{5}{c}{BLEU} & \multicolumn{5}{c}{COMET} \\
 & \textbf{S} & \textbf{T} & \textbf{I} & \textbf{R} & \textbf{All} & \textbf{S} & \textbf{T} & \textbf{I} & \textbf{R} & \textbf{All} \\
\hline
\rowcolor{gray!20} \multicolumn{11}{c}{Open-source LVLMs}\\
BLIP2-Opt-6.7B & 0.00 & 0.00 & 0.00 & 0.00 & 0.00 & 46.67 & 45.25 & 46.31 & 46.09 & 45.97 \\
InstructBLIP-Vicuna-7B & 0.00 & 0.00 & 0.00 & 0.00 & 0.00 & 46.67 & 45.25 & 46.31 & 46.09 & 45.97 \\
mPLUG-Owl3-7B-241101 & 0.00 & 0.32 & 2.79 & 0.00 & 0.41 & 46.76 & 45.40 & 47.97 & 46.11 & 46.24 \\
LLaVA1.5-7B & 0.24 & 0.21 & 2.26 & 0.00 & 0.38 & 46.70 & 45.57 & 47.09 & 46.09 & 46.18 \\
LLaVA1.5-13B & 1.72 & 2.61 & 0.63 & 0.84 & 1.71 & 48.26 & 47.35 & 46.79 & 46.39 & 47.32 \\
LLaVA-Next-Interleave-Qwen-7B & 0.00 & 1.11 & 0.89 & 0.00 & 0.50 & 46.65 & 46.04 & 47.04 & 46.12 & 46.33 \\
LLaVA-OneVision-Qwen2-7B & 6.36 & 2.57 & 4.31 & 4.14 & 4.22 & 51.15 & 47.41 & 49.57 & 48.96 & 49.08 \\
DeepSeek-VL2-Small-16B & 0.58 & 0.16 & 0.52 & 0.28 & 0.35 & 47.14 & 45.36 & 46.80 & 46.22 & 46.23 \\
DeepSeek-VL2-27B & 0.82 & 0.16 & 0.54 & 0.00 & 0.35 & 47.24 & 45.36 & 46.88 & 46.11 & 46.24 \\
Llama3.2-11B-Vision-Instruct & 0.90 & 0.00 & 4.10 & 0.00 & 0.69 & 47.59 & 45.32 & 48.56 & 46.46 & 46.59 \\
MiniCPM-Llama3-V2.5-8B & 2.56 & 0.24 & 0.00 & 0.27 & 0.88 & 49.41 & 45.41 & 46.31 & 46.28 & 46.86 \\
MiniCPM-V2.6-8B & 5.59 & 5.46 & 5.40 & 12.80 & 7.29 & 51.16 & 49.00 & 49.62 & 54.47 & 51.02 \\
Qwen2-VL-7B-Instruct & 8.07 & 5.52 & 14.04 & 0.23 & 5.85 & 53.07 & 49.52 & 56.56 & 46.28 & 50.48 \\
Qwen2.5-VL-7B-Instruct & \textbf{16.08} & \textbf{12.04} & \textbf{18.94} & 7.72 & \textbf{12.86} & 57.48 & 54.34 & \textbf{59.38} & 51.40 & \textbf{55.05} \\
InternVL2-8B & 15.81 & 0.69 & 3.90 & 9.00 & 7.37 & \textbf{58.01} & 45.66 & 48.22 & 52.28 & 51.07 \\
InternVL2.5-8B & 11.26 & 6.84 & 1.28 & \textbf{10.39} & 8.38 & 54.31 & 50.94 & 47.72 & \textbf{52.49} & 51.94 \\
InternVL2.5-8B-MPO & 13.23 & 10.91 & 14.03 & 8.44 & 11.30 & 55.60 & \textbf{54.80} & 53.58 & 50.45 & 53.84 \\
\hline
\rowcolor{gray!20} \multicolumn{11}{c}{Closed-source LVLMs}\\
GPT-4o-2024-11-20 & 32.82 & 23.23 & 36.82 & 22.50 & 27.22 & 67.66 & 60.30 & 68.69 & 60.74 & 63.39 \\
GPT-4.5-preview-2025-02-27 & \textbf{46.19} & 33.54 & \textbf{60.92} & \textbf{54.82} & \textbf{45.25} & 77.65 & 69.34 & \textbf{81.79} & 81.40 & \textbf{75.98} \\
Moonshot-V1-8K & 30.03 & 20.77 & 6.84 & 24.93 & 22.96 & 66.57 & 60.01 & 51.41 & 62.31 & 61.53 \\
Qwen-VL-Max-2024-11-19 & 43.11 & 25.66 & 49.30 & 48.23 & 38.65 & 76.32 & 64.86 & 76.55 & 78.51 & 72.70 \\
Qwen-VL-Max-2025-01-25 & 41.67 & 34.11 & 54.42 & 53.81 & 43.23 & 75.29 & 71.45 & 79.76 & \textbf{81.74} & 75.94 \\
Gemini-1.5-Flash & 45.79  & 39.95  & 44.90  & 35.41  & 41.02  & \textbf{79.07}  & 74.12  & 72.94  & 69.11  & 74.18 \\
Gemini-2.0-Flash & 44.85 & \textbf{42.53} & 43.58 & 44.06 & 43.68 & 76.20 & \textbf{74.14} & 75.17 & 75.00 & 75.05 \\
\hline
\end{tabular}
\end{adjustbox}
\end{center}
\vspace{-0.2cm}
\caption{The BLEU and COMET scores of the English-Chinese menu translation.} \label{evadata_en2zh}
\end{table*}

\begin{table*}[ht]
\begin{center}
\begin{adjustbox}{width=1.45\columnwidth,height=3.7cm,center}
\begin{tabular}{lcccccccccc}
\hline
\multirow{2}{*}{CN2EN} & \multicolumn{5}{c}{BLEU} & \multicolumn{5}{c}{COMET} \\
 & \textbf{S} & \textbf{T} & \textbf{I} & \textbf{R} & \textbf{All} & \textbf{S} & \textbf{T} & \textbf{I} & \textbf{R} & \textbf{All} \\
\hline
\rowcolor{gray!20} \multicolumn{11}{c}{Open-source LVLMs}\\
BLIP2-Opt-6.7B & 0.00 & 0.00 & 0.00 & 0.00 & 0.00 & 39.12 & 40.08 & 42.42 & 37.63 & 39.26 \\
InstructBLIP-Vicuna-7B & 0.00 & 0.00 & 0.00 & 0.00 & 0.00 & 39.12 & 40.08 & 42.42 & 37.63 & 39.26 \\
mPLUG-Owl3-7B-241101 & 0.05 & 0.00 & 0.50 & 0.24 & 0.13 & 39.19 & 40.08 & 42.53 & 37.80 & 39.34 \\
LLaVA1.5-7B & 0.00 & 0.14 & 0.00 & 0.00 & 0.04 & 39.12 & 40.14 & 42.42 & 37.63 & 39.28 \\
LLaVA1.5-13B & 0.00 & 0.29 & 0.00 & 0.15 & 0.13 & 39.12 & 40.22 & 42.42 & 37.70 & 39.32 \\
LLaVA-Next-Interleave-Qwen-7B & 0.00 & 0.07 & 0.50 & 0.22 & 0.13 & 39.12 & 40.16 & 42.62 & 37.86 & 39.37 \\
LLaVA-OneVision-Qwen2-7B & 0.28 & 0.49 & 0.99 & 0.30 & 0.41 & 39.46 & 40.35 & 42.86 & 37.83 & 39.56 \\
DeepSeek-VL2-Small-16B & 0.04 & 0.00 & 0.00 & 0.00 & 0.01 & 39.23 & 40.08 & 42.42 & 37.63 & 39.27 \\
DeepSeek-VL2-27B & 0.00 & 0.00 & 0.00 & 0.00 & 0.00 & 39.12 & 40.08 & 42.42 & 37.63 & 39.26 \\
Llama3.2-11B-Vision-Instruct & 0.32 & 0.25 & 0.50 & 0.27 & 0.30 & 39.50 & 40.29 & 42.62 & 38.09 & 39.60 \\
MiniCPM-Llama3-V2.5-8B & 0.42 & 2.65 & 0.66 & 0.48 & 1.13 & 39.84 & 42.45 & 43.17 & 38.50 & 40.53 \\
MiniCPM-V2.6-8B & 6.95 & 8.46 & 13.26 & 7.98 & 8.25 & 48.78 & 48.10 & 53.68 & 48.18 & 48.83 \\
Qwen2-VL-7B-Instruct & 5.17 & 1.52 & 6.06 & 0.88 & 2.90 & 44.14 & 42.07 & 47.60 & 38.98 & 42.32 \\
Qwen2.5-VL-7B-Instruct & 5.98 & 2.51 & 6.63 & 5.09 & 4.73 & 44.18 & 44.31 & 46.55 & 43.99 & 44.37 \\
InternVL2-8B & 8.22 & 4.62 & \textbf{15.07} & \textbf{9.21} & 8.02 & \textbf{49.18} & 45.53 & \textbf{56.19} & \textbf{48.98} & 48.63 \\
InternVL2.5-8B & 4.59 & 3.10 & 14.96 & 1.97 & 4.28 & 44.25 & 43.41 & 53.88 & 40.77 & 43.83 \\
InternVL2.5-8B-MPO & \textbf{8.68} & \textbf{11.12} & 12.57 & 8.12 & \textbf{9.59} & 47.72 & \textbf{52.95} & 52.24 & 46.34 & \textbf{49.29} \\
\hline
\rowcolor{gray!20} \multicolumn{11}{c}{Closed-source LVLMs}\\
GPT-4o-2024-11-20 & 12.87 & 10.11 & 14.54 & 9.44 & 11.19 & 52.30 & 50.46 & 54.90 & 49.04 & 51.03 \\
GPT-4.5-preview-2025-02-27 & 24.74 & 25.41 & 23.72 & 18.70 & 23.11 & 64.09 & 64.65 & 61.18 & 58.81 & 62.48 \\
Moonshot-V1-8K & 25.86 & 25.12 & 30.14 & 19.62 & 24.20 & 65.08 & 64.23 & 68.64 & 60.05 & 63.68 \\
Qwen-VL-Max-2024-11-19 & 21.22 & 19.36 & 22.59 & 16.13 & 19.31 & 59.14 & 58.66 & 60.95 & 55.90 & 58.21 \\
Qwen-VL-Max-2025-01-25 & 17.67 & 19.46 & 23.83 & 19.63 & 19.63 & 58.67 & 60.91 & 60.53 & 60.69 & 60.69 \\
Gemini-1.5-Flash & 12.31  & 7.55  & 28.94  & 14.53  & 12.96  & 54.62  & 48.03  & 66.65  & 52.92  & 53.19 \\
Gemini-2.0-Flash & \textbf{31.58} & \textbf{32.49} & \textbf{40.45} & \textbf{26.82} & \textbf{31.25} & \textbf{71.58} & \textbf{71.20} & \textbf{76.24} & \textbf{69.33} & \textbf{71.22} \\
\hline
\end{tabular}
\end{adjustbox}
\end{center}
\vspace{-0.2cm}
\caption{The BLEU and COMET scores of the Chinese-English menu translation.} \label{evadata_zh2en}
\vspace{-0.5cm}
\end{table*}


\paragraph{English-Chinese Translation Evaluation Result.} The evaluation results of the English-Chinese (EN2CN) menu translation are shown in Table \ref{evadata_en2zh}. The English-to-Chinese menu translation capabilities of open-source LVLMs are generally weak, with extremely limited upper bounds (Qwen2.5-VL-7B-Instruct achieves a BLEU score of 12.86 and a COMET score of 55.05). In contrast, the EN2CN menu translation capabilities of closed-source LVLMs are relatively stronger (GPT-4.5-preview-2025-02-27 achieves a BLEU score of 45.25 and a COMET score of 75.98).


\paragraph{Chinese-English Translation Evaluation Result.} The evaluation results of the Chinese-English (CN2EN) menu translation are shown in Table \ref{evadata_zh2en}. Although the CN2EN menu translation capability of open-source LVLMs is still weaker than that of closed-source LVLMs, the open-source LVLM and closed-source LVLM with the strongest EN2CN menu translation capability do not have the strongest Chinese-English menu translation capability. The open-source LVLM and closed-source LVLM with the strongest CN2EN menu translation capability are InternVL2.5-8B-MPO and Gemini-2.0-Flash, respectively.


Overall, although the most advanced closed-source models, such as GPT-4.5-preview-2025-02-27 and Gemini-2.0-flash, outperform the smaller open-source models on MOTBench, some open-source models, such as Qwen2.5-VL-7B-Instruct and InternVL2.5-8B-MPO, have also demonstrated strong competitiveness. Moreover, optimized versions of VLLMs from the same series generally show improved OCR and translation capabilities compared to their previous versions. This not only proves the effectiveness of their optimization strategies but also further validates the reliability of MOTBench.


\section{Consistency Analysis Between Human and Automated Evaluations}


To verify the consistency between the automatic evaluation and human evaluation, we take human evaluation as the ground truth and calculate the accuracy of automatic evaluation results of menu OCR task and automatic extraction translation results of menu translation task.

\subsection{Accuracy Analysis of MOTBench Automatic evaluation Method on Menu OCR Task}

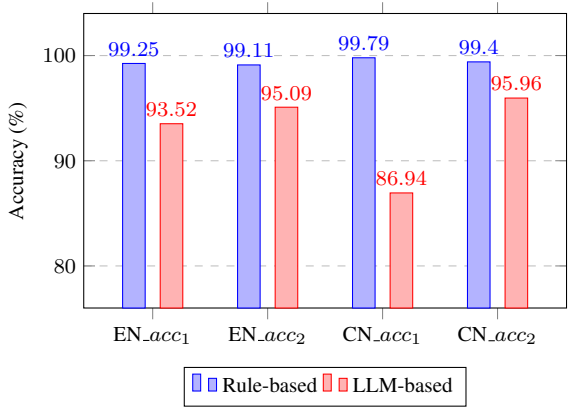
\begin{figure}[ht]
\centering
\hspace{-0.7cm}
\begin{tikzpicture}
    \begin{axis}[
        ybar=0.2cm,
        symbolic x coords={EN\_$acc_1$,EN\_$acc_2$,CN\_$acc_1$,CN\_$acc_2$},
        xtick=data,
        ylabel={Accuracy (\%)},
        legend style={at={(0.5,-0.2)}, anchor=north, legend columns=0.9},
        enlargelimits=0.2,
        width=8cm,
        height=5.5cm,
        bar width=0.3cm,
        ymajorgrids=true,
        grid style=dashed,
        nodes near coords,
        nodes near coords align={vertical}, 
        ymin=80, ymax=100,
    ]
        \footnotesize \addplot coordinates {(EN\_$acc_1$,99.25) (EN\_$acc_2$,99.11) (CN\_$acc_1$,99.79) (CN\_$acc_2$,99.40)};
        \addplot coordinates {(EN\_$acc_1$,93.52) (EN\_$acc_2$,95.09) (CN\_$acc_1$,86.94) (CN\_$acc_2$,95.96)};
        \legend{Rule-based, LLM-based};
    \end{axis}
\end{tikzpicture}
\caption{With human evaluation serving as the ground truth, the accuracy of both rule-based and LLM-based automatic evaluation results for the menu OCR task are measured. Here, accuracy is defined as the consistency between the automatic evaluation and human evaluation.}
\label{fig:ocr}
\end{figure}

Taking human evaluation as the ground truth, we analyze the accuracy of rule-based and LLM\textsuperscript{\ref{llama3.1}}-based automatic evaluation results in the menu OCR task. Specifically, we select the menu OCR results generated by MiniCPM-V2.6-8B and calculate the accuracy  ($acc_1$ and $acc_2$) of rule-based and LLM-based automatic evaluation results based on human evaluation. As shown in Figure \ref{fig:ocr}, the accuracy of the rule-based evaluation results exceeds 99\%, while the accuracy of the LLM-based automatic evaluation results is significantly lower than that of the rule-based evaluation results. The performance gap is most pronounced for Chinese dish evaluations (CN\_$acc_1$), where the difference reaches 12.85\%. This is the main reason we adopt the rule-based automatic evaluation method for the menu OCR task.

\subsection{Accuracy Analysis of Automated evaluation Method on Menu Translation Task}


In the menu translation task, the automatic evaluation method consists of two steps: automatically extracting dish translation results and calculating translation-related automatic evaluation metrics. We compute the translation-related automatic evaluation metrics based on BLEU and COMET, which are two mainstream automatic evaluation metrics for translation-related tasks. Their reliability has been thoroughly validated, so we do not conduct further analysis here. Instead, we focus on analyzing the accuracy of automatically extracted dish translation results.

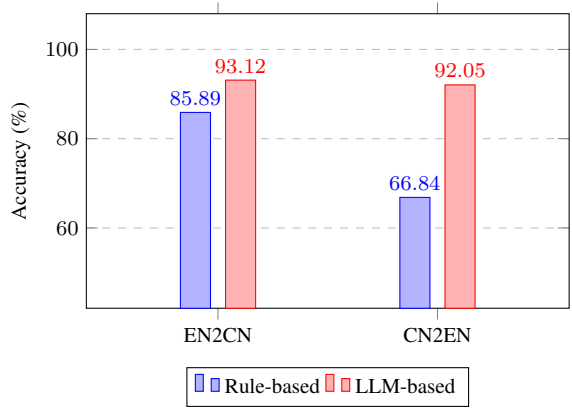
\begin{figure}[ht]
\centering
\hspace{-0.7cm}
\begin{tikzpicture}
    \begin{axis}[
        ybar=0.2cm,
        symbolic x coords={EN2CN,CN2EN},
        xtick=data,
        ylabel={Accuracy (\%)},
        legend style={at={(0.5,-0.2)}, anchor=north, legend columns=0.9},
        enlargelimits=0.6,
        width=8cm,
        height=5.5cm,
        bar width=0.4cm,
        ymajorgrids=true,
        grid style=dashed,
        nodes near coords,
        nodes near coords align={vertical}, 
        ymin=60, ymax=90,
    ]
        \footnotesize \addplot coordinates {(EN2CN,85.89) (CN2EN,66.84)};
        \addplot coordinates {(EN2CN,93.12) (CN2EN,92.05)};
        \legend{Rule-based, LLM-based};
    \end{axis}
\end{tikzpicture}
\caption{With human evaluation serving as the ground truth, the accuracy of both rule-based and LLM-based methods for automatically extracting dish translations in the menu translation task are measured. Here, accuracy is defined as the consistency between the automatic extraction and human evaluation.}
\label{fig:mt}
\end{figure}

Specifically, we select the menu translation results generated by MiniCPM-V2.6-8B and then calculate the accuracy of rule-based and LLM\textsuperscript{\ref{llama3.1}}-based dish translation extraction methods using human evaluation as the ground truth. As shown in Figure \ref{fig:mt}, the accuracy of the LLM-based extraction method is as high as 92\% or more (93.12\% for EN2CN and 92.05\% for CN2EN), while the accuracy of the rule-based extraction method is significantly lower. This is the primary reason we adopt the LLM-based extraction method for dish translation results in the menu translation task.

\section{Conclusion}


This paper introduces MOTBench, a benchmark specifically designed to evaluate the application of LVLMs in understanding complex layout documents, particularly in the field of menu translation. MOTBench comprehensively evaluate the visual understanding and language processing capabilities of models by requiring LVLMs to accurately recognize and translate each dish on the menu along with its price and unit items. Experimental results show that the automatic evaluation outcomes are highly consistent with professional human evaluation. By evaluating a series of state-of-the-art LVLMs, this paper identifies the strengths and weaknesses of these models, providing valuable insights for the future development of LVLMs.


{\small

\bibliographystyle{ieee_fullname}
}

\end{document}